\documentclass{article}
\pdfoutput=1

\usepackage{arxiv}
\usepackage{amsmath}
\usepackage{algorithm}
\usepackage{algcompatible}
\usepackage{adjustbox}
\usepackage{amsfonts}
\usepackage{amssymb}
\usepackage{amsbsy}
\usepackage[T1]{fontenc}
\usepackage{textcomp}
\usepackage{graphicx}
\usepackage{listings}
\usepackage{multirow}
\usepackage[normalem]{ulem}
\usepackage{color}
\definecolor{ruby}{rgb}{0.6,0,0.3}

\usepackage{epstopdf}
\usepackage[caption=false]{subfig}

\usepackage{multirow}
\usepackage{xparse}
\ExplSyntaxOn
\NewDocumentCommand{\smallcaps}{m}
 {
  \tl_set:Nn \l_tmpa_tl { #1 }
  \regex_replace_all:nnN
   { ([0-9]+) } 
   { \c{resizedigit}\cB\{ \1 \cE\} } 
   \l_tmpa_tl
  \textsc{ \tl_use:N \l_tmpa_tl }
 }

\ExplSyntaxOff

\usepackage{tikz}
\newcommand*\circled[1]{\tikz[baseline=(char.base)]{
            \footnotesize\node[shape=circle,draw,inner sep=1pt] (char) {#1};}}

\usepackage[hidelinks]{hyperref}
\usepackage{url}

\graphicspath{{./}}
\newcommand{\ignore}[1]{}
\newcommand{\revisit}[1]{}

\title{Unsupervised ore/waste classification on open-cut mine faces using close-range hyperspectral data}
\shorttitle{Unsupervised ore/waste classification on mine faces using hyperspectral data}

\date{} 

\author{
  This article is published in \textit{Geoscience Frontiers} (volume 14, issue 4). It is available online at\\
  {\color{blue}\url{https://doi.org/10.1016/j.gsf.2023.101562}} under the \href{http://creativecommons.org/licenses/by-nc-nd/4.0/}{CC BY-NC-ND} license.\vspace{2mm}\\
  {\footnotesize\circled{c} 2023 China University of Geosciences (Beijing) and}\\
  {\footnotesize\,\, Peking University. Journal hosted by Elsevier B.V.}\\ \\
  \textbf{Lloyd~Windrim}, \textbf{Arman~Melkumyan}\thanks{Author for correspondence (post-acceptance): arman.melkumyan@sydney.edu.au}, \textbf{Richard~J.~Murphy},\\ \textbf{Anna~Chlingaryan} and \textbf{Raymond~Leung}\thanks{Corresponding author (pre-acceptance): raymond.leung@sydney.edu.au}\vspace{2mm} \\
  Rio Tinto Centre for Mine Automation\\
  Australian Centre for Field Robotics (ACFR)\\
  The University of Sydney, Australia
}

\renewcommand{\headeright}{Windrim et al.}
\renewcommand{\undertitle}{Accepted article}

\begin{document}
\maketitle

\begin{abstract}
The remote mapping of minerals and discrimination of ore and waste on surfaces are important tasks for geological applications such as those in mining. Such tasks have become possible using ground-based, close-range hyperspectral sensors which can remotely measure the reflectance properties of the environment with high spatial and spectral resolution. However, autonomous mapping of mineral spectra measured on an open-cut mine face remains a challenging problem due to the subtleness of differences in spectral absorption features between mineral and rock classes as well as variability in the illumination of the scene. An additional layer of difficulty arises when there is no annotated data available to train a supervised learning algorithm. A pipeline for unsupervised mapping of spectra on a mine face is proposed which draws from several recent advances in the hyperspectral machine learning literature. The proposed pipeline brings together unsupervised and self-supervised algorithms in a unified system to map minerals on a mine face without the need for human-annotated training data. The pipeline is evaluated with a hyperspectral image dataset of an open-cut mine face comprising mineral ore martite and non-mineralised shale. The combined system is shown to produce a superior map to its constituent algorithms, and the consistency of its mapping capability is demonstrated using data acquired at two different times of day.
\end{abstract}

\vspace{-2mm}

\vspace{5mm}\noindent\textbf{Highlights}
\begin{itemize}
\item Self-supervised learning for mine face hyperspectral image ore/waste classification
\item Spectral angle stacked auto-encoder (RSA-SAE) learns illumination-invariant features
\item Pre-trained VNIR composite network plus transfer learning reduces convergence time
\item Training CNN with spectral relighting augmented data further increases the F1 score
\item Resultant network classifies pixel spectra with precision/recall around 97.2\%-99.4\%
\end{itemize}

\newpage\section{Introduction}\label{sec:intro}
As sensor technology has advanced, so too has research into techniques to support a fully autonomous mining workflow \cite{Leung2022overview}. One such sensor, a hyperspectral camera, is capable of imaging the environment so that the absorption properties of scene constituents can be measured remotely at a high spatial and spectral resolution. Each channel of a hyperspectral image captures light at a narrow range of wavelengths within the visible, near-infrared or short-wave infrared portions of the electromagnetic spectrum. Thus, a single pixel in a hyperspectral image can be used to measure the reflectance characteristics of an encompassed material to a higher spectral resolution and beyond the visible range sensed by most other systems (e.g. a conventional RGB camera) \cite{Schowengerdt2007}. 

\begin{figure*}[!t]
{\includegraphics[width=1\textwidth, clip=true,trim= 0 0 0 0]{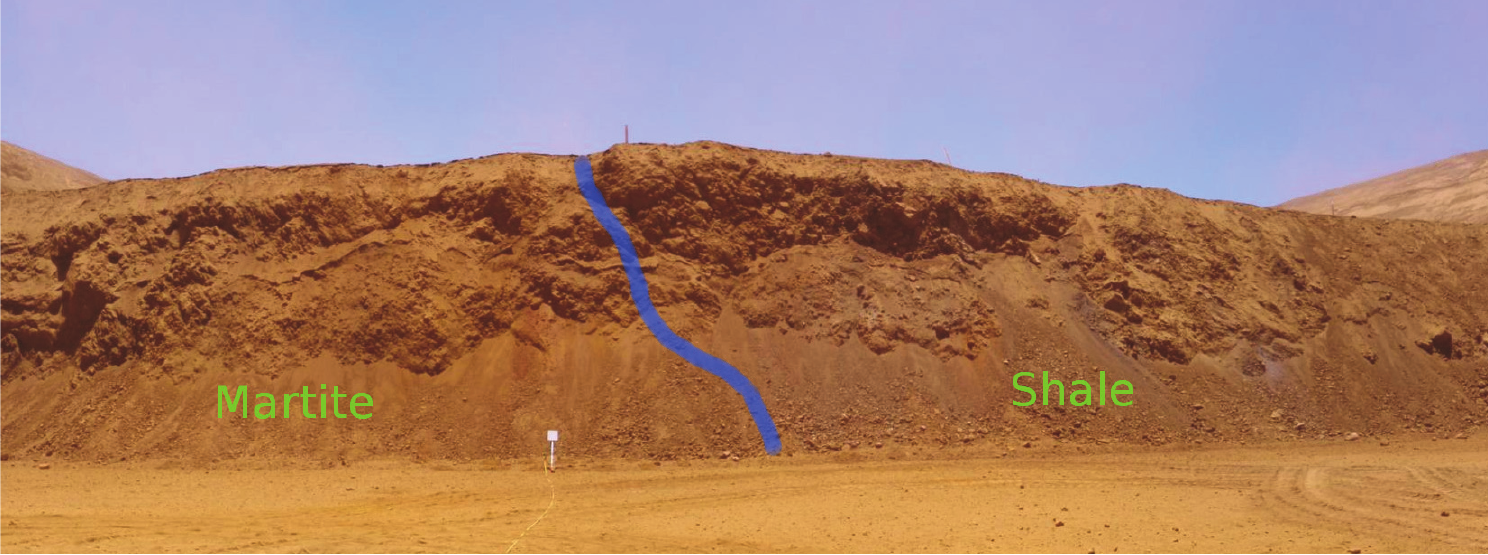}}
\caption{A mine face containing martite and shale \cite{Murphy2012}. These two materials are difficult to differentiate to the naked eye and most RGB cameras. }
\label{fig:mine face}
\end{figure*}

In geological applications such as open-pit mining, autonomous systems must discriminate between minerals of varying composition and waste that appear similarly to conventional RGB cameras. It is often subtle differences in the reflectance spectra that allow these distinctions to be made and these differences can only be measured with a high-resolution hyperspectral camera. For example, in Murphy (2012) zones of ore-bearing minerals goethite and martite are distinguished from non-ore-bearing shale zones by finely-resolved features in the spectral domain using hyperspectral cameras \cite{Murphy2012}. Using a conventional RGB camera, such a distinction would be difficult (see Figure~\ref{fig:mine face}). 

Autonomous mining systems that use hyperspectral imagery to map the material distribution on the surface of open-cut mine faces usually rely on some form of labelled dataset to either train a supervised classification algorithm \cite{Windrim2016a,Schneider2011} or build a spectral library to which field spectra can be matched in order to identify the material \cite{Schneider2009,Murphy2012}. Collecting and labelling geological hyperspectral data to build the dataset can be done in different ways, where each is challenging for its own reasons. One approach is to collect rock samples from the field and take them back to a laboratory where spectrometer data is recorded and analysed in order to assign a class label. The spectrometer data with the label is then added to a database of labelled spectra. The problems with this approach are that it is time-consuming and expensive as it requires special expertise and resources. Additionally, spectrometer data from a laboratory is usually acquired by illuminating samples with artificial lighting and hence does not exhibit the variability that field data does (\cite{Hecker2008}). To compensate for this, illumination and other effects must be artificially simulated within the data---which itself is a challenging task.

An alternate and often more representative method of collecting and labelling geological hyperspectral data is to label data directly from field imagery, whose spectra exhibit variability due to the illumination and other sources. The challenge with this approach is in the difficulty of identifying materials in mine face imagery through visual inspection of composite images, which have similar colour and texture (e.g. Figure~\ref{fig:mine face}). Therefore, pixel spectra are typically inspected individually using specialised software such as ENVI \footnote{https://www.l3harrisgeospatial.com/Software-Technology/ENVI} to assign labels, which is time-consuming and hard because of the variability in the field data. 

A combination of the two approaches can be used whereby the locations of rock samples acquired from the field are tagged such that they can be located in the imagery. Once identified using spectrometer analysis, the regions at the tagged locations in the imagery can be labelled efficiently. The disadvantage of this approach is that it requires people to be in the field to collect samples and do the tagging. Mine sites are dangerous locations for people as there are usually large moving vehicles and hazardous activities going on such as blasting.

The challenges of labelling hyperspectral data have motivated research into techniques which are either unsupervised or only use limited amounts of labelled training data. Unsupervised methods do not use labelled data for training machine learning models. Examples are feature extraction techniques developed for hyperspectral \cite{Demarchi2014b,Licciardi2009,Romero2016a,Romero2014}, endmember extraction \cite{tu2000unsupervised,close2012using,winter1999n}, segmentation \cite{Acito2003,Grana2009a} and clustering methods such as $k$-means. The problem with these approaches is that they often are not robust to the variability of hyperspectral data---particularly on a mine face, which affects the mapping performance. The complex geometry of a mine face results in self-shadowing, variations in the angle that the illumination source strikes the surface, indirect illumination from nearby reflections, and a range of other phenomena \cite{Windrim2018,Murphy2012}. Together, these effects produce per-pixel variations in the incident illumination and intensity.

Some of these problems were tackled in the research of Windrim (2016, 2018, 2019a) \cite{Windrim2016,Windrim2018,Windrim2019a} which incorporated functions from the hyperspectral literature (e.g. Spectral Angle \cite{Yuhas1992}, Spectral Information Divergence \cite{Chang2000}) with some illumination invariant properties into the loss functions of autoencoders. Windrim (2018) \cite{Windrim2018} also used a strategy of training the autoencoder to map artificially shadowed versions of spectra to their original sunlit equivalents to encode shadow invariance into the learnt features. This method, called the Relit Spectral Angle - Stacked Autoencoder (RSA-SAE), proved effective at learning shadow invariant features. However, because it relied on no prior knowledge of the scene, it required randomly sampling the parameters of an atmospheric modeller to generate a range of candidate terrestrial sunlight and diffuse skylight spectra necessary to simulate the shadow.

Windrim (2016a) \cite{Windrim2016a} proposed a similar idea with supervised classification models which could be trained on a limited set of training samples. Relighting is the process of simulating the spectral appearance of a region under different illumination and geometrical conditions that are not encompassed by the training set. Using spectral relighting augmentation, supervised models could be trained on a small number of sunlight samples and have the models generalised to other illumination conditions within the scene. Unlike Windrim (2018) \cite{Windrim2018}, these models estimated the ratio between the terrestrial sunlight and diffuse skylight spectra from the image, which alleviates the need for sampling. The additional benefit of supervised models such as these is that they can leverage pre-trained models that have been exposed to non-task specific labelled data as in \cite{Windrim2018a}, which demonstrated how a model pre-trained on airborne agricultural hyperspectral data could be fine-tuned on close-range hyperspectral data captured from a mine face. This, in conjunction with spectral relighting augmentation, reduces the requirement for a large labelled training dataset from a mine face in order to train a mineral mapping algorithm. Despite this, some labelled training data from the mine face is still required, which is dangerous for the reasons described above.

In this paper we investigate how the ideas proposed in three key papers: RSA-SAE \cite{Windrim2018}, spectral relighting augmentation \cite{Windrim2016a} and pre-training/transfer learning on larger labelled datasets \cite{Windrim2018a}, can be used together in a unified pipeline to address the challenging problem of autonomous mineral mapping on an open-cut mine face, without any labelled data. Section \ref{sec:method} describes the study site and dataset used for experiments, as well as the method of the pipeline. Experimental results are presented in section \ref{sec:Results} and discussed in section \ref{sec:discuss}, before conclusions are drawn in section \ref{sec:concl}.

The technical novelty of this work may be summarised in two parts. First, illumination-invariant features are extracted from mine face hyperspectral images via unsupervised learning. An RSA stacked autoencoder is used to produce low-dimensional representations that are consistent within a given class and largely independent of shadows and lighting conditions. The objective is to obtain confident class representatives via clustering in this latent space where the confounding effects due to lighting/shadows are minimised. Second, material classification is achieved via a self-supervised process using a hyperspectral CNN, whereby a spectral relighting augmentation strategy is used to increased diversity in the training data obtained from the first stage. This is combined with transfer learning to reduce the number of epochs required to reach convergence. The efficacy of the proposed approach is demonstrated, wherein transfer learning uses a VNIR composite network pre-trained on data that is unrelated to mineral mapping.

\subsection{Related Work}\label{into:rel_work}
Several studies have developed and applied supervised machine learning techniques to close-range hyperspectral imagery to map geological materials on vertical mine faces. Murphy (2012) \cite{Murphy2012} carried out extensive experiments to compare the mapping performance of the Spectral Angle Mapper (SAM) and Support Vector Machines (SVMs) for ferric and non-ferric minerals. Due to its greater robustness to the adverse effects of illumination, SAM was shown to be more effective than the SVM for this task. Schneider (2011, 2011a) \cite{Schneider2011,Schneider2011a} used an Observation Angle Dependant (OAD) covariance function to instil illumination invariant properties onto Gaussian Processes (GPs), which were shown to outperform SAM and conventional GPs on mineral mapping applications. Chlingaryan (2016) \cite{Chlingaryan2016} showed that Multi-Task Gaussian Processes (MTGPs), previously used for spectral unmixing on a mine face \cite{Uezato2014}, could also out-perform SAM. Several more recent studies have demonstrated the effectiveness of convolutional neural networks for mapping minerals on mine faces \cite{Windrim2016a,Windrim2018a}. For an overview of various deep learning approaches which have been proposed for hyperspectral classification, readers are referred to \cite{Audebert2019deep} where different architectures, design choices, and issues relating to spatial and spectral resolution are discussed.

Beyond machine learning, there are other interesting research with demonstrated application to close-range geological mapping, such as the registration of hyperspectral data with 3D models. It has been shown that LiDAR-derived terrain information can be used to correct geometric distortions in the hyperspectral data \cite{Kurz2008,Buckley2012}, and also to enhance geological interpretation \cite{Krupnik2016,Snyder2016,Monteiro2013}. Similarly, Salehi (2018), Lorenz (2018) and Kirsch (2018, 2019) \cite{Salehi2018,Lorenz2018,Kirsch2018,Kirsch2019} proposed workflows that integrate spectral products with 3D photogrammetric terrain data, demonstrating the advantages of terrestrial hyperspectral sensing for mapping vertical geological outcrops in extreme regions not observable by air or spaceborne sensors. Ghamisi (2018) \cite{Ghamisi2018} surveyed and analysed other classification techniques, such as those based on mathematical morphology, Markov random fields, tree partitioning and sparse representation. These techniques are well suited to airborne hyperspectral images which are generally not affected by shadows that appear on rough surfaces of a vertical mine face.

Several works have explored the use of characteristic absorption features to map the geology of vertical surfaces. Using the depth of a characteristic absorption feature, outcrops of carbonatite \cite{Boesche2015a} and zones enriched by neodymium \cite{Boesche2015} were mapped. Investigations were conducted into the effects of illumination on clay absorption features \cite{Murphy2014}  and ferric iron crystal field absorption features \cite{Murphy2014a} for geological mapping. These studies provide important insights as well as new capabilities that can be leveraged by machine learning algorithms. Krupnik (2019) \cite{Krupnik2019} have done a more detailed literature review of studies involving close-range hyperspectral sensing for mining applications. In a project funded by the Minerals Research Institute of WA, Austin et al. \cite{Austin2019UQ} used hyperspectral imaging to determine ore grade at a mine face in real-time at the point of excavation \cite{Choros2022}. This work demonstrated a capability which is considered essential in the development of autonomous mining excavators.

Self-supervised learning is scarcely used in geological hyperspectral imaging applications. However, hyperspectral deep learning is a rapidly evolving field. Recently, Liu and Han \cite{Liu2022deep} used deep self-supervised learning to reconstruct high-resolution hyperspectral images (HR-HS) by merging low-resolution hyperspectral images (LR-HS) with high-resolution RGB images. Their architecture contained a generative backbone which is conceptually similar to ours, in that both try to capture the prior underlying structure in the latent HR-HS image. A point of difference is that Liu and Han also considered spatial structures and image texture whereas our proposal focuses on spectral reflectance and achieving illumination invariance. In Liu et al.\,\cite{Liu2022es2fl}, EfficientNet-80 was used as the backbone to model input samples following a spatial-spectral approach. A key part of their proposal was constraining the cross-correlation matrix of different distortions of the same sample to the identity matrix, which enabled the model to extract latent features of homogeneous and heterogeneous samples, coalescing the former and separating the latter in a self-supervised manner. Their feature learning and classification was based on an ensemble learning strategy which jointly utilised spatial context information at different scales and feature information in different bands. In Song et al.\,\cite{Song2022self}, a self-supervised branch appeared in a semi-supervised residual network (SSRNet). Discriminative features were learned by performing two tasks: masked bands reconstruction and spectral order forecast. Santiago et al.\,\cite{Santiago2022deep} employed self-supervised cluster assignments between sets of contiguous bands to learn semantically meaningful representations. As the suitability of these techniques is yet to be established across different application domains, this study seeks to consolidate the prior works of Windrim et al.\,\cite{Windrim2016a,Windrim2018,Windrim2018a} and integrate targeted approaches that work well for hyperspectral mineral mapping.

\section{Material and Methods}\label{sec:method}
A pipeline is constructed to cluster (and hence map) each pixel of a hyperspectral image of an open-cut mine when there is no labelled training data or knowledge of classes available. In this problem, the only prior knowledge available is the number of classes in the scene. Everything else, including the names and whereabouts of the classes in the image, is unknown. The pipeline is evaluated with a close-range hyperspectral image of an iron ore mine face.

\subsection{Study Area and Data Capture}\label{sec:method:dataset}
Experiments are conducted on a dataset from a West Angelas mine in the Hamersley Province, in the Pilbara Region of Western Australia. The data was compiled by Murphy et al.\,\cite{Murphy2012}. A Visual Near Infrared (VNIR) image is captured using the Specim AISA Eagle hyperspectral camera, a line scanner with $220$ channels in the range $400-970$ nm and a spectral resolution of 2 nm. The image is captured 30 metres from the mine face, so the spatial resolution of the image is approximately 6 cm per image pixel. The camera was mounted on a rotating stage to acquire the image. Two images were captured of the face: one at 11:30 and one at 13:30, such that the illumination effects change between the two images. A 30cm x 30cm calibration panel of known reflectance (99\%-reflective Spectralon) is placed in the scene for normalising the pixel spectra to apparent reflectance, and the images are spatially registered to correct any misalignments. 

\begin{figure*}[!htb]
{\includegraphics[width=1\textwidth, clip=true,trim= 0 0 0 0]{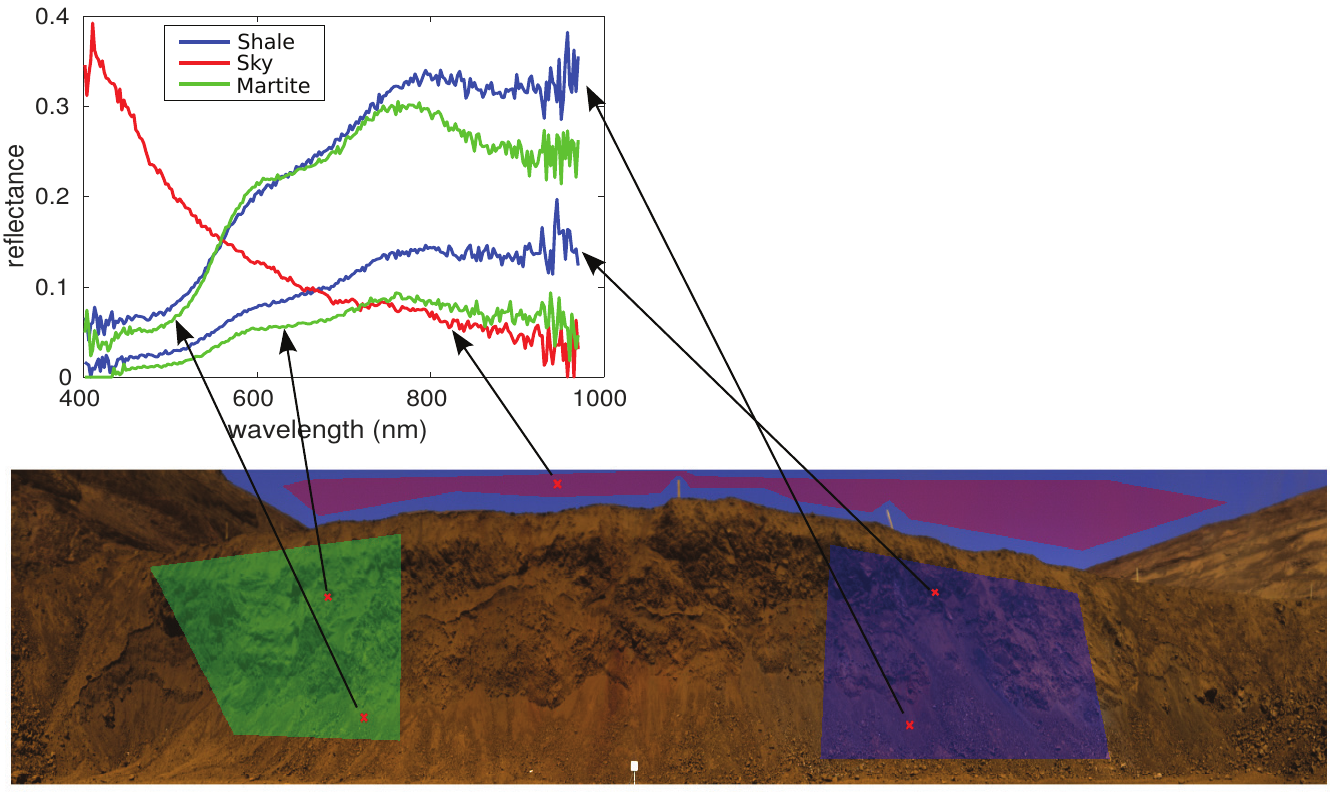}}
\caption[The 11:30 mine face image with ground truth areas highlighted.]{The 11:30 mine face image with ground truth areas highlighted (colour scheme: green-martite, blue-shale, red-sky). An example spectrum from each class is shown. The similarity of the shale and martite classes coupled with the within-class variability of the spectra across the scene makes this a difficult clustering problem.}
\label{fig:caseStudy_gt}
\end{figure*}

The face contains two main geological classes: martite and shale, as well as some sky in the background (Figure~\ref{fig:caseStudy_gt}). Hence, there are three predominant classes in the image. A geologist has identified a rough boundary dividing the two classes on the mine face through X-ray diffraction analysis and laboratory spectroscopy of various samples collected in the field. There are actually two types of shale; however, there is no ground truth information for separating these two classes, so they are considered to be a single class for these experiments. As evidenced in the spectra of Figure~\ref{fig:caseStudy_gt}, separating the sky from the geological classes is a trivial task, although, separating the geological classes is difficult due to their similarity coupled with illumination-induced variability in spectra from across the scene. 

The spectral data in the image of the mining scene has variability due to the interaction of incident light with its complex topology. There are many shaded regions throughout the image. Classifying or clustering spectra with significant variability is difficult. Even if there was prior knowledge of which classes were in the scene, the variability would still make it difficult to accurately classify the hyperspectral image (e.g. using Spectral Angle Mapper or SAM with a reference library).

\subsection{Pipeline}\label{sec:method:pipeline}
There are two main parts to this pipeline. Firstly, an unsupervised process is used to categorise the data and extract some class representative points. This process only extracts points of high confidence, and hence does not capture the variability of the class. But it removes the need for any spectral libraries or reference data, and doesn't require knowledge of the labels of the classes in the scene. Once high confidence class representatives have been extracted from the scene, a self-supervised process is trained to predict the class association of all pixels in the image. Within this process, the variability in the data, due to illumination and the scene's complex topology, is accounted for. This is essentially clustering the data a second time, but this time, more accurately. Note that whilst supervised algorithms are used, the entire process requires no labelled data. The data to train the supervised algorithm comes from the unsupervised process (hence it is self-supervised).

\begin{figure}[!htb]
\centering
{\includegraphics[width=0.9\textwidth, clip=true,trim= 0 0 0 0]{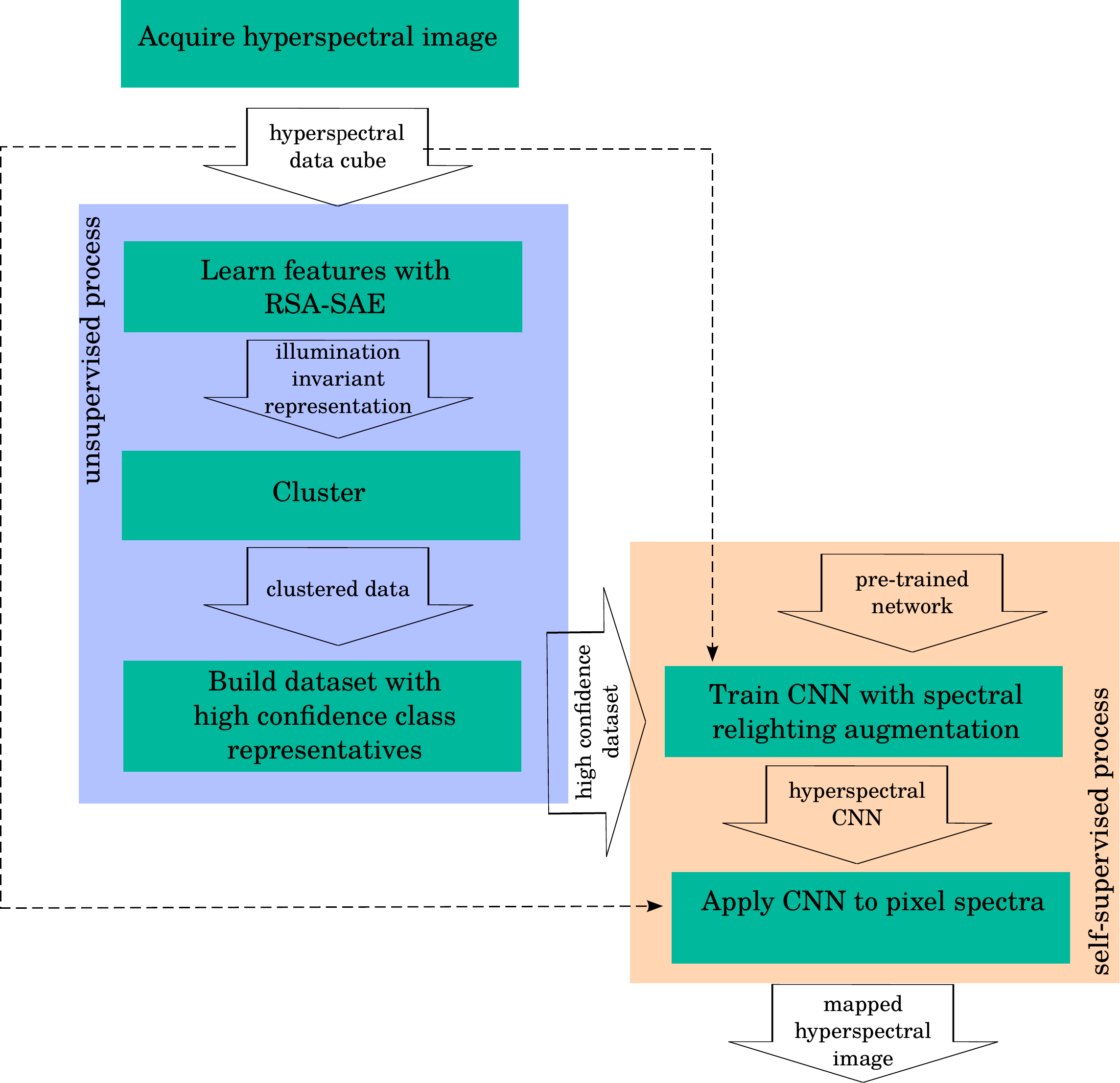}}
\caption[A flowchart summarising the pipeline for clustering.]{A flowchart summarising the pipeline for clustering the pixels of a hyperspectral image without any prior labelled data.}
\label{fig:caseStudy_pipeline}
\end{figure}

The result of the pipeline is a function which can produce a per-pixel thematic map of the distribution of the classes in a hyperspectral image. The pipeline is summarised in Figure~\ref{fig:caseStudy_pipeline}.

\begin{figure}[!htb]
\centering
{\includegraphics[width=0.8\textwidth, clip=true,trim= 0 0 0 0]{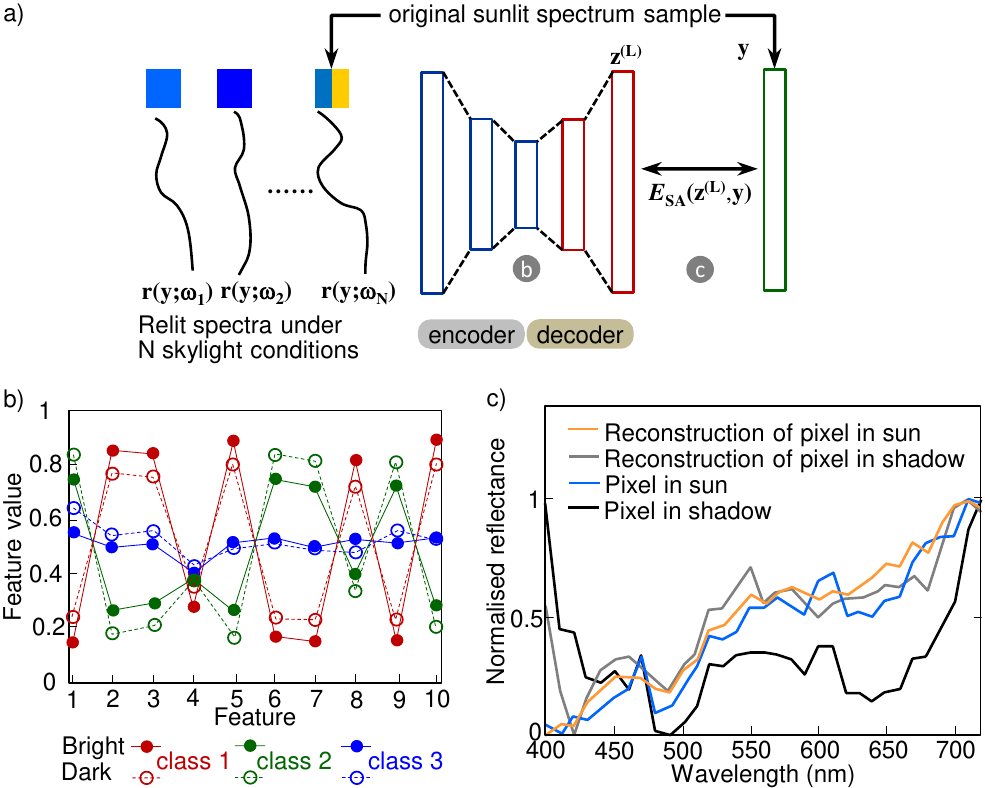}}
\caption[Framework for training the RSA-SAE network]{(a) A framework for training the RSA-SAE network. (b) Encoded features for spectra from three different classes under bright and dark lighting conditions. (c) Comparison of spectrum and reconstruction for pixel spectra sampled in sunlit and shadowed regions.}
\label{fig:rsa-sae}
\end{figure}

The RSA-SAE component is designed to overcome challenges posed by self-shadowing and variable lighting conditions which usually degrade classification performance. The goal is to encode each material consistently under sunlit and shadowed conditions. A framework for training the RSA-SAE network is depicted in Figure~\ref{fig:rsa-sae}a. The colour patches indicate the illumination type, with blue and yellow representing skylight and sunlight respectively. A pixel in a shadowed region is illuminated only by skylight, hence its patch is coloured entirely in blue. Different shades of blue represent different atmospheric conditions. The network is trained to reconstruct the original sunlit spectrum (the green bar $\mathbf{y}$) from the input spectrum after it has been relit in shadow using a relighting equation ($r$) and candidate atmosphere ($\omega_n$). Training optimises the weights in the network layers to minimise the spectral angle cost function $E_\text{SA}(\mathbf{z}^{(L)},\mathbf{y})$ described in \cite{Windrim2018}. Once training completes, the encoder is expected to map real shadowed and sunlit spectra from the same class to roughly the same values in the code layer. This provides a similar, low-dimensional representation for data within the same class (see Figure~\ref{fig:rsa-sae}b). Ideally, the network captures the essential features in each respective class, which enables the decoder to reconstruct spectra that are alike (see yellow and grey curves in Figure~\ref{fig:rsa-sae}c) irrespective of shadows and lighting conditions.

\subsubsection{Unsupervised Process}\label{sec:unsupervised-process}
An RSA-SAE network \cite{Windrim2018} is trained on the hyperspectral data from the 11:30 image. Note that it could be trained on the 13:30 image, or data from both images, but the 11:30 image is selected to show that the pipeline generalises to new images captured with different illumination conditions. In doing this, a mapping is learnt from the high dimensional hyperspectral reflectance image to a low dimensional, illumination invariant image. The same network architecture, meta parameters and training procedure as \cite{Windrim2018} is used. The input data has 220 features (one for each spectral channel). The encoder has three fully-connected (i.e. dense) layers of size 100, 50 and 30 neurons. The decoder is symmetric, having three fully-connected layers of size 50, 100 and 220. Hence, a 30 dimensional illumination invariant representation is learnt at the deepest hidden layer of the network. Each layer of the network is first pre-trained on all 289×1443 (i.e. the spatial resolution of the image) pixel spectra for 1000 epochs using the Cosine Spectral Angle-SAE \cite{Windrim2016} before an end-to-end fine-tuning process with the RSA-SAE is done using 5000 sampled pixel spectra. See \cite{Windrim2018} for more details about the RSA-SAE method.

Once the spectral data has been mapped to an illumination invariant feature space with the RSA-SAE, it is clustered using the $k$-means algorithm, searching for three clusters (it is assumed that prior knowledge is available that there are three classes in the image). Then, the dataset of high confidence class representatives for training the classifier is built by automatically extracting the 200 points closest to each of the three cluster centroids. The distance is measured using the Euclidean distance. These 200 points have the highest confidence of belonging to each of the classes in the scene. Hence, the dataset has 600 points in total. 

\subsubsection{Self-Supervised Process}\label{sec:self-supervised-process}
\begin{figure*}[!t]
{\includegraphics[width=1\textwidth]{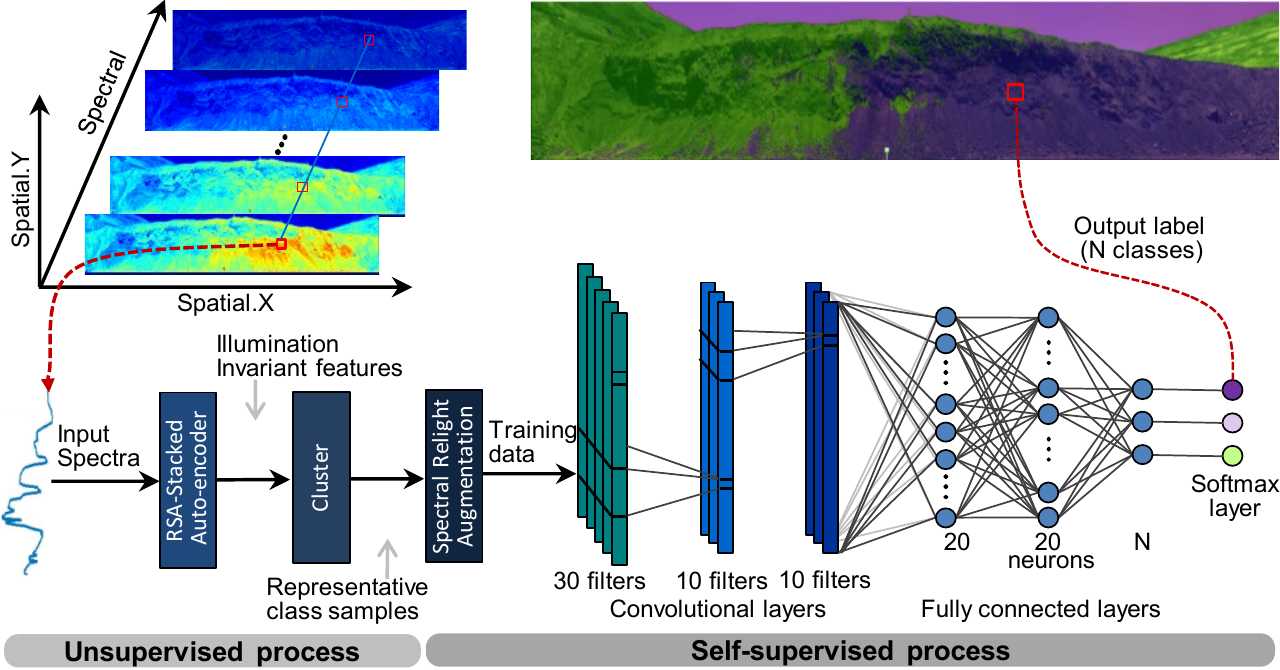}}
\caption[Network architecture]{Simplified network architecture. The auto-encoder described in Sec.~\ref{sec:unsupervised-process} contains three fully-connected (FC) layers with 100, 50 and 30 neurons. It converts a spectrum with 220 channels to a 30 dimensional illumination invariant representation. The decoder is symmetric and contains 3 FC layers of size 50, 100 and 220. Confident class samples are extracted by clustering and subject to spectral relighting augmentation to produce 600 training points covering three classes (ore, waste and sky). This data is split 9:1 for training and validation. The architecture of the CNN is described in Sec.~\ref{sec:self-supervised-process}.}
\label{fig:network-architecture}
\end{figure*}

The parameters of a hyperspectral Convolutional Neural Network (CNN) are initialised from a network trained on different labelled VNIR data as in \cite{Windrim2018a}, using the same composite VNIR dataset reported in that paper. The pre-training dataset is a combination of spectra from the Pavia University and Salinas datasets, with classes like asphalt, meadows and bitumen. There are no mining-related classes in these datasets---their purpose is for the network to learn useful and generic spectral features to reduce the training load for the mining task. The hyperspectral CNN has a similar architecture to the networks reported in \cite{Windrim2018a} and \cite{Windrim2016a}. This particular network has three one-dimensional convolutional and three fully-connected layers, with batch normalisation and ReLU activation layers. The convolutional layers have filters of size 30, 10 and 10 respectively. The fully-connected layers have size 20, 20 and 3 neurons. The network architecture is depicted in Figure~\ref{fig:network-architecture}.

The dataset automatically extracted from the high confidence points via the unsupervised process is divided into training and validation data, with 180 training points per class and 20 validation points per class. The validation samples are to ensure the model does not overfit. To train the CNN, batches of data are extracted from the training partition and augmented using spectral relighting as described in \cite{Windrim2016a}, which expands the size of the batch by many times. The augmented batch is then interpolated to the spectral wavelengths used by the pre-trained network, and pre-processed by offsetting the spectra so that its mean is zero, before being input into the CNN. The CNN is trained for 200 epochs with stochastic gradient descent \cite{Bottou2010}. Once the CNN is trained, it can be applied to the remaining pixels in the image and to the 13:30 image to produce thematic classification maps. 

\subsection{Experimental Setup}
The different stages of the pipeline are evaluated qualitatively to ensure each component functions as expected. The output of the entire process is then evaluated quantitatively with some labelled pixels using the F1 score (the harmonic mean between precision and recall). An ablation experiment is done whereby the key components of the self-supervised process are removed from the pipeline and the output assessed quantitatively to determine their impact on the process.

The pipeline was implemented on a 64-bit computer with an Intel Core i7-4770 CPU @ 3.40GHz × 8 processor and GeForce GTX 760/PCIe/SSE2 graphics card. All CNNs were trained using the \texttt{matconvnet-1.0-beta20} software package \cite{Vedaldi2015} and the experiments were run in MATLAB. The library was compiled with GPU support using the CUDA toolkit.

\section{Results}\label{sec:Results}
\begin{figure*}[!t]
\centering
\includegraphics[width=0.95\textwidth]{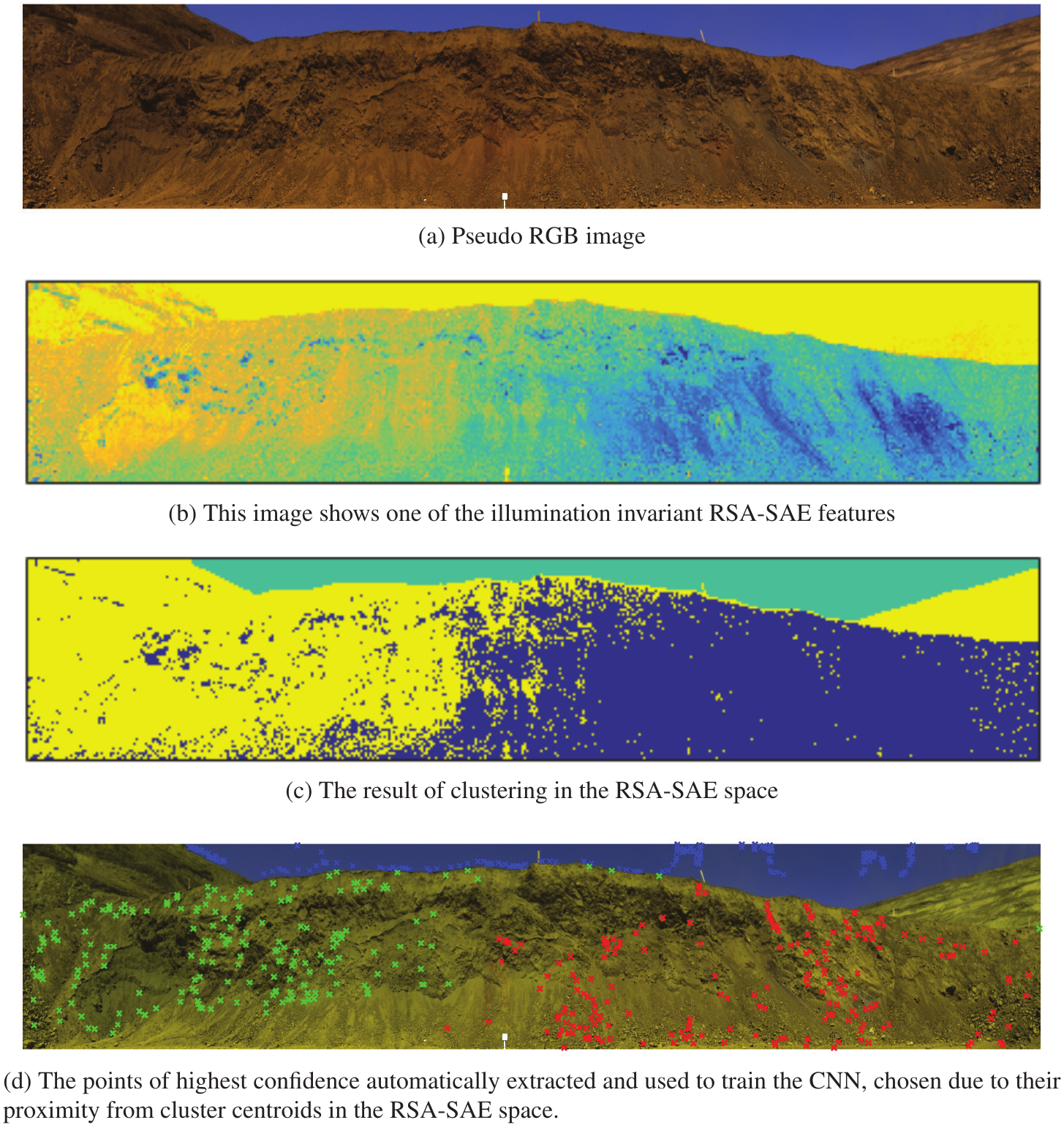}
\caption[Results from the different steps of the unsupervised process.]{Results from the different steps of the unsupervised process, in the extraction of the annotated dataset for training the CNN. The results are from the 11:30 mining image.}
\label{fig:result_caseStudy_pipeline}
\end{figure*}

\begin{figure*}[!t]
\centering
\includegraphics[width=0.95\textwidth]{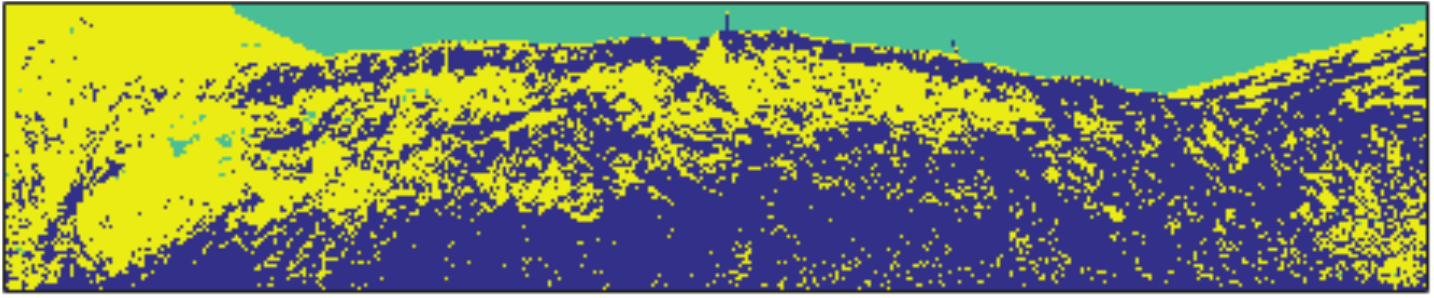}
\caption{The result of clustering in the original reflectance space.}
\label{fig:result_caseStudy_cluster_raw}
\end{figure*}

The results at different stages of the pipeline for the unsupervised process are shown in Figure~\ref{fig:result_caseStudy_pipeline}. The invariant image (Figure~\ref{fig:result_caseStudy_pipeline}b) visualises the representation of the hyperspectral image with one of the RSA-SAE features. It was easier to see how the materials were distributed across the mine face in the invariant image in comparison to the pseudo RGB image, where everything was uniform in colour and texture. The shadows in the invariant image were also far less prominent. The clustered image (Figure~\ref{fig:result_caseStudy_pipeline}c) aligned well with the ground truth information (Figure~\ref{fig:caseStudy_gt}). The result was far less affected by the illumination than if the data was clustered in the original reflectance space (Figure~\ref{fig:result_caseStudy_cluster_raw}). The high-confidence points that were automatically identified by being closest to the cluster centroids (Figure~\ref{fig:result_caseStudy_pipeline}d) appeared to belong to the correct class (according to the boundary identified by geologists). An interesting observation is that these points, which were used to train the self-supervised CNN, are generally well illuminated. They rarely coincide with the different classes under shadow.

\begin{figure*}[!t]
\centering
\includegraphics[width=0.95\textwidth]{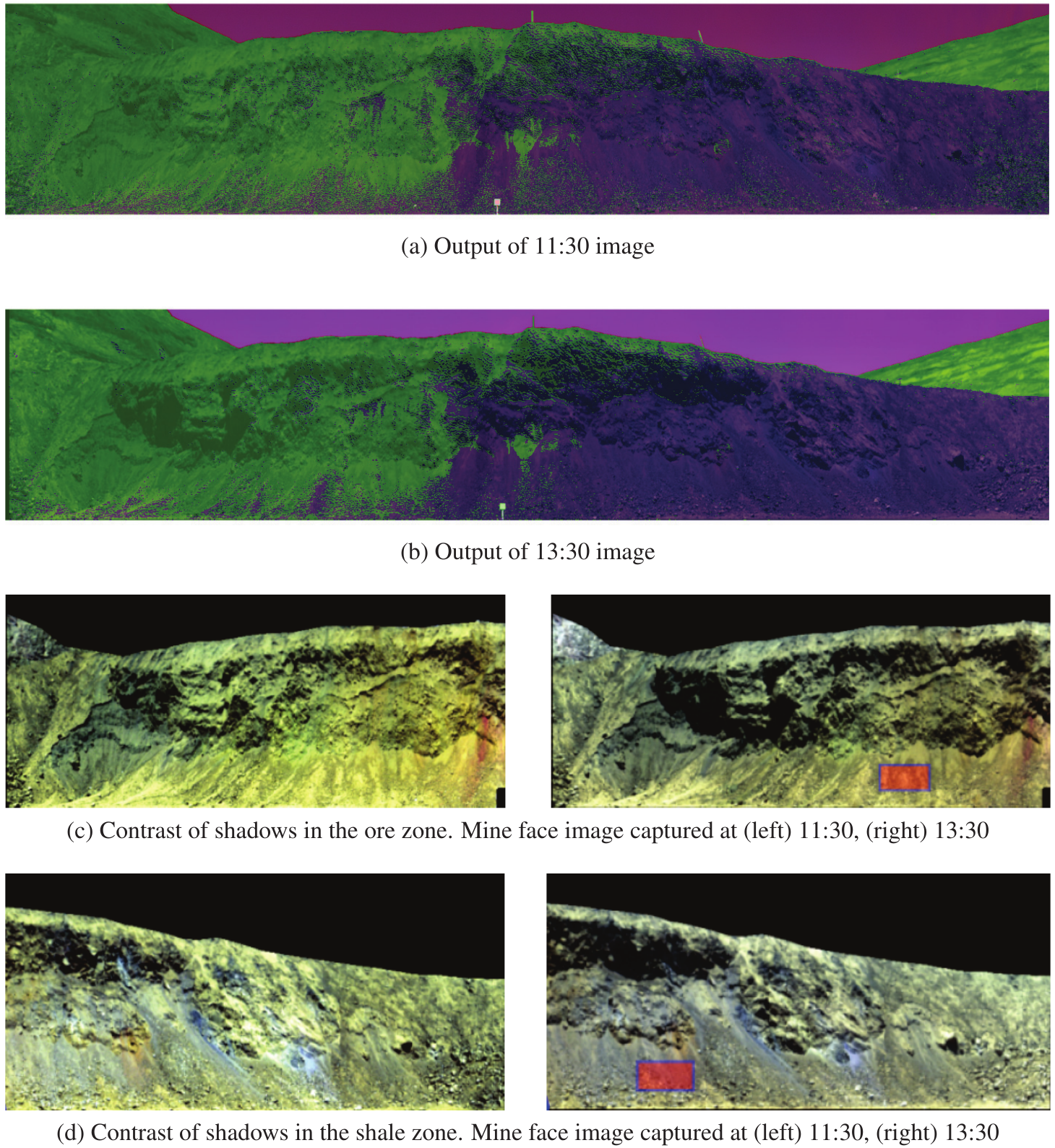}
\caption[Images of the mine face captured at different times of the day but assigned categories with the same CNN.]{Images of the mine face captured at different times of the day but assigned categories with the same CNN, thus showing the temporal generality of the pipeline. Green represents martite prediction, purple represents shale prediction, and pink represents sky prediction.}
\label{fig:result_caseStudy_class}
\end{figure*}

\begin{figure*}[!htb]
\centering
\includegraphics[width=0.7\textwidth]{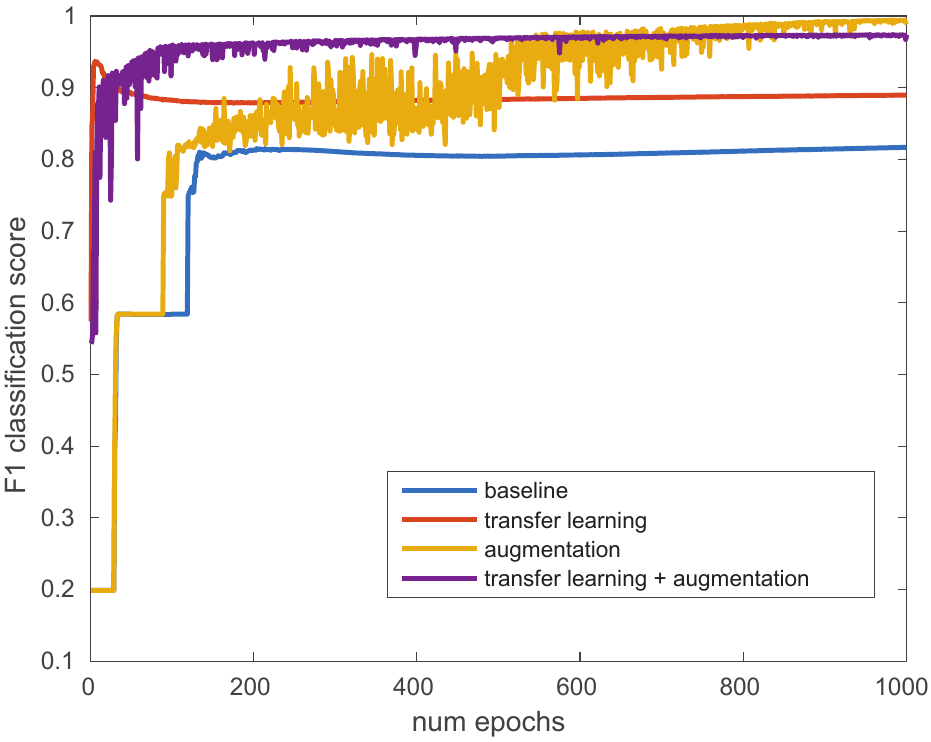}
\caption[Change in the F1 score for the different elements of the self-supervised process.]{A comparison of the change in the F1 score as the optimisation progresses for the different elements of the self-supervised process.}
\label{fig:result_caseStudy_comparison}
\end{figure*}

The training dataset that was extracted using the unsupervised process was then used to train a CNN in the self-supervised component of the pipeline. A visualisation of applying the trained classifiers to the 11:30 and 13:30 images is shown in Figure~\ref{fig:result_caseStudy_class} and the progression in classification score during optimisation is shown in Figure~\ref{fig:result_caseStudy_comparison} for four different training strategies, using a test set of about 120,000 labelled samples from the 11:30 image for evaluation. The four training strategies included a transfer learning approach where the self-supervised CNN classifier was pre-trained from the VNIR composite network (with no augmentation), an augmentation strategy where the parameters were trained from scratch (no pre-training) with spectral relighting augmentation, and a combined approach of spectral relighting augmentation and pre-training the network from the VNIR composite. A baseline approach with no pre-training or data augmentation was also compared against for benchmarking. 

The visualisation of the mining images with the CNN clustering result overlaid (Figure~\ref{fig:result_caseStudy_class}) showed good correspondence to the ground truth information and the geological boundary. There was also insignificant change in the clustered pixels across the two times. Most of the regions which had a significant change in the illumination conditions between 11:30 and 13:30 remained in the same predicted cluster. 

The result in Figure~\ref{fig:result_caseStudy_comparison} compares the individual significance of each of the key elements of the self-supervised process, that is, the transfer learning and spectral relighting augmentation. When used on its own, in comparison to the baseline CNN, the transfer learning via initialisation with a pre-trained network reduced the number of epochs required to reach convergence, and the F1 score at convergence was also better. When the spectral relighting augmentation was used on its own, it took more epochs for the optimisation to converge than the baseline, but the F1 score it converged at was nearly perfect (99.4\%), significantly higher than the baseline (81.7\%). When using both transfer learning and spectral relighting augmentation, the CNN converged to a very good result with relatively few epochs. The F1 score (97.2\%) was higher than both the baseline and transfer learning only approaches (with F1 scores of 81.7\% and 89.9\%, respectively), and was just under the spectral relighting augmentation-only result. The number of epochs required for convergence was comparable to the transfer learning-only approach, and far fewer than the number of epochs required for the spectral relighting augmentation-only result.

\section{Discussion}\label{sec:discuss}
Converting the hyperspectral image into the illumination invariant RSA-SAE space was critical for the success of the pipeline. Figure~\ref{fig:result_caseStudy_pipeline} examines the overall effectiveness of the RSA-SAE. The effects of self-shadowing which make the task of ore/waste discrimination difficult can be seen in Figure~\ref{fig:result_caseStudy_pipeline}a. In Figure~\ref{fig:result_caseStudy_pipeline}b, the mine face is rendered using the values for one of the learned features. This validation step reveals two striking observations. First, the colour pixels show spatial separation that is broadly consistent with mineralogy---with orange-yellow corresponding to martite (ore) and green-blue corresponding to shale (waste). Second, this result is largely independent of the location of shadows. In contrast, when the data was clustered in the original reflectance space (Figure~\ref{fig:result_caseStudy_cluster_raw}), the shadows had a negative effect on the clusters, making them inaccurate representations of the class distributions on the mine face. This is demonstrated by the correlation between the shape of the clusters in Figure~\ref{fig:result_caseStudy_cluster_raw} and the distribution of shadow on the image in Figure~\ref{fig:result_caseStudy_pipeline}a. A possible solution is to search for extra clusters such that the shadows effects become confined to dedicated clusters. The problem with this is that it is essentially creating extra classes. If the self-supervised learning were to be conducted using these extra clusters, it would be very hard to automatically determine which clusters belonged to the same class or which clusters were from shadows. However, by clustering in the illumination invariant space, the number of clusters can be set to the number of classes in the scene, and the shadows have almost no impact on the clustering accuracy, with shadowed and sunlit data from corresponding classes being correctly clustered together (Figure~\ref{fig:result_caseStudy_pipeline}c). The lack of correlation between the image clustered in the illumination invariant space and the distribution of shadows demonstrates the effectiveness of the RSA-SAE for finding illumination invariant feature representations.

The points of high confidence that were close to the cluster centroids were all in sunlit regions (Figure~\ref{fig:result_caseStudy_pipeline}d), which is to be expected, as they were most similar to the mean spectra of that cluster. This means that they do not capture the variability needed to train a robust classifier. At this stage of the pipeline, there was a small amount of labelled training data available that captured a limited amount of the variability in the scene. This is why the spectral relighting augmentation methods for training are necessary to classify the entire scene correctly.

The results (Figure~\ref{fig:result_caseStudy_class}) show that when spectral relighting augmentation and transfer learning were used with the CNN, the pixels were correctly grouped despite the training data not capturing the variability. By transferring knowledge from the airborne VNIR datasets, the classification accuracy on the mining data was improved and the convergence time was reduced (Figure~\ref{fig:result_caseStudy_comparison}). The F1 classification score at convergence for the network using purely transfer learning was better than the baseline approach. This is impressive, because in order to do the transfer learning, the data had to be reduced in spectral range from $401-970$ nm to $430-860$ nm, which is the spectral range of the VNIR pre-trained network. This means that despite having less information from the mining dataset, the network was able to produce better classification accuracies by leveraging information from other datasets, highlighting the advantage of the transfer learning. Although convergence was significantly faster, the purely transfer learning CNNs F1 score at convergence was not as high as the purely augmented CNNs score. This was expected, because it was only trained on the high confidence points extracted from the image via the unsupervised process. The dataset did not capture the variability, and the spectral relighting augmentation added the missing variability needed to train the classifier. The best result was achieved by combining the two techniques (transfer learning and spectral relighting augmentation) together. By doing this, the merits of both approaches were achieved (fast convergence and high accuracy). The F1 classification score at convergence (97.2\%) was slightly lower for the combined approach than when the augmentation was used on its own (99.4\%). This is to be expected because the combined approach also required that the wavelength range be reduced to $430-860$ nm, as it used the pre-trained VNIR network for initialisation. This could be changed by training a new pre-trained network which uses the full VNIR spectrum, and transferring knowledge from that dataset instead. 

The F1 score fluctuated more for the methods that used spectral relighting augmentation (Figure~\ref{fig:result_caseStudy_comparison}). This is because in each batch that was used to train the CNN, the parameters for the relighting were being randomly sampled, meaning the same batch of data from  two different epochs could be very different. But as the CNN came closer to converging, the fluctuations reduced in size. 

In Figure~\ref{fig:result_caseStudy_class}, the boundary between the predicted classes deviates slightly from the ground truth geological boundary. This could be due to a non-illumination related source of variability in the spectra and poor representation in the self-supervised training samples for these challenging examples. Despite this inaccuracy, the predicted boundary is consistent across the two times. The scree is also a source of error in the prediction.

\begin{table*}[htbp]
  \centering
  \caption{Runtime for different stages of the pipeline.}
    \begin{tabular}{ccc}\\ \hline
    \textbf{Stage} & \multicolumn{1}{l}{\textbf{Processor}} & \multicolumn{1}{l}{\textbf{Time}} \\ \hline
  pre-train RSA-SAE & CPU & 60m 31s \\
  train RSA-SAE & CPU &  14m 51s\\
  apply RSA-SAE & CPU & $<$1s \\
  clustering & CPU & 1s \\
  train CNN & GPU & 23m 43s \\
  apply CNN & GPU & 10s \\ \hline
    \end{tabular}%
  \label{tab:result_caseStudy_runtime}%
\end{table*}%

The runtimes in Table~\ref{tab:result_caseStudy_runtime} were measured using 1000 epochs for training the autoencoder and CNN. The time to train the CNN with transfer learning and spectral relighting augmentation was 1.42 seconds per epoch (about 24 minutes to train it for 1000 epochs), but, due to the transfer learning, at about 250 epochs, the convergence reached within one percent of its final value. Thus, a good classifier could be trained in about six minutes. Once trained, to forward propagate every pixel in the image through the CNN only took 10 seconds, which was significantly shorter than the training time. Additionally, running new images of this size through the network would take a similarly shorter amount of time. The longest stage of the pipeline was the pre-training of the RSA-SAE. In this process, each layer was trained individually for 1000 epochs. It could be possible to skip this step all together if a generic network was trained on lots of data and used to pre-train RSA-SAE networks for any new image (similar to how the self-supervised CNN is pre-trained). It is also possible to speed up the training of the RSA-SAE by using a GPU instead of the CPU.

In terms of impact, this work has demonstrated progress on three fronts. First, unsupervised feature learning is viable in the context of ground-based close-range hyperspectral mineral mapping---for ore/waste discrimination, labelled data is not required. Second, approximate invariance to shadows and lighting conditions is achievable, this represents a significant hurdle that the RSA-SAE has helped overcome. Reducing the confounding effects due to shadows and changes in illumination fundamentally improves classification performance, but it also expands the deployment window in practice to longer hours and a wider range of environmental conditions. Finally, transfer learning using a VNIR composite model pre-trained on non-mining data suggests the approach may be used in different settings, for instance, in copper-zinc deposits, as well as iron ore open-cut mines.

\section{Conclusion and Future Work}\label{sec:concl}
Given the occupational hazards present at a mine site and the difficulty in labelling hyperspectral data, there is a need for unsupervised algorithms that do not require labelled data. This work develops a pipeline comprising three key ideas from recent research. The relit spectral-angle stacked auto-encoder (RSA-SAE), pre-training\,/\,transfer learning, and spectral relighting augmentation work together to enable mineral mapping without labelled data. Clustering in an illumination-invariant feature space is used to extract training data for a convolutional neural network. This approach is combined with transfer learning and data augmentation to achieve better results. The pipeline is evaluated on two close-range hyperspectral images of a vertical iron ore mine face, captured at different times of day. The pipeline outperformed its individual components in terms of accuracy and speed.

The proposed method allows workers to be removed from hazardous parts of the mining process and improve efficiency. Besides crush and blast injuries, there are also health risks associated with inhalation of airborne fibrous minerals which are prominent in certain banded iron formations in the Pilbara. Potential detection of these hazardous material prior to drilling can help minimise human exposure, and alert site geologists to activate risk mitigation and management plans. For instance, a registered engineer may provide close supervision, oversee the containment strategy associated with blasting, and subsequent load-haul operations to designated fibrous waste disposal sites.

Future work will explore alternatives to the requirement of prior knowledge of the number of classes. Illumination models which are more complex and incorporate other indirect illumination effects will also be explored. Finally, the impact of alternative sampling techniques (e.g. random sampling) for self-selecting the training spectra will be investigated.

\section{Acknowledgements}
This work has been supported by the Rio Tinto Centre for Mine Automation and the Australian Centre for Field Robotics, University of Sydney.

\bibliographystyle{unsrt}
\bibliography{references.bib}

\end{document}